\documentclass[10pt,twocolumn,letterpaper]{article}

\usepackage[table]{xcolor}
\usepackage[utf8]{inputenc}

\usepackage{iccv}
\usepackage{times}
\usepackage{graphicx}
\usepackage{amsmath}
\usepackage{amssymb}

\usepackage[inline]{enumitem}
\usepackage{array}

\usepackage{tikz}
\usepackage{pgfplots}
\input{tikz-pgfplots.settings}

\def\thicknesslineplot{ultra thick}

\definecolor{gtcolor}{rgb}{0,0,0}
\definecolor{ourcolor}{rgb}{0,0.4,0.6}
\definecolor{ourcolor2}{rgb}{0.4,0.2,0.4}
\definecolor{dingcolor}{rgb}{0.8,0.1,0.3}
\definecolor{kukelovacolor}{rgb}{0,0.6,0}
\definecolor{jiangcolor}{rgb}{0,0.7,0.7}
\definecolor{inliercolor}{rgb}{0.2,1,0.1}
\definecolor{outliercolor}{rgb}{1,0,0}
\definecolor{smallcolor1}{rgb}{0.,0,0}
\definecolor{smallcolor2}{rgb}{0,0,0}

\usepackage{standalone}
\pgfplotsset{
    table/search path={graphs},
}

\usepackage{bm}
\renewcommand{\vec}[1]{\bm{#1}}
\providecommand{\mat}[1]{\bm{#1}}

\usepackage{mathtools}

\newcommand{\R}{\ensuremath{\mathbb{R}}}
\newcommand{\te}[1]{\text{#1}}
\DeclarePairedDelimiterX{\norm}[1]{\lVert}{\rVert}{#1}
\DeclarePairedDelimiterX{\abs}[1]{\lvert}{\rvert}{#1}
\newcommand{\fr}[2]{\frac{#1}{#2}}

\newcommand{\T}{T}
\DeclareMathOperator{\tr}{tr}

\DeclareMathOperator{\diag}{diag}
\makeatletter
\newcommand{\vast}{\bBigg@{4}}
\newcommand{\Vast}{\bBigg@{5}}
\makeatother

\usepackage[all]{foreign}
\renewcommand{\etal}{\emph{et~al.}}
\renewcommand{\ie}{\emph{i.e.}}
\renewcommand{\eg}{\emph{e.g.}}

\newcommand{\Ey}{\ensuremath{\mat{E}_{y}}}
\newcommand{\Ry}{\ensuremath{\mat{R}_{y}}}
\newcommand{\skewmat}[1]{\ensuremath{\left[#1\right]_{\times}}}
\graphicspath{{images/}}

\usepackage{multirow}
\usepackage{booktabs}

\usepackage[pagebackref=true,breaklinks=true,letterpaper=true,colorlinks,bookmarks=false]{hyperref}
\usepackage[capitalize,noabbrev]{cleveref}

\iccvfinalcopy 

\def\iccvPaperID{3947} 

\newcommand{\hackspace}[1]{\vspace{#1}}

\begin{document}
%
\title{Trust Your IMU: Consequences of Ignoring the IMU Drift\hackspace{-3mm}}
\def\ww{1mm}
\author{Marcus Valtonen \"Ornhag$^1$ \hspace{\ww} Patrik Persson$^{1}$ \hspace{\ww} M{\aa}rten Wadenb{\"a}ck$^2$ \hspace{\ww} Kalle {\AA}str{\"o}m$^1$ \hspace{\ww} Anders Heyden$^1$\\[1mm]
	\begin{minipage}[c]{0.4\textwidth}
		\centering
		${}^1$Centre for Mathematical Sciences\\
		Lund University
	\end{minipage}
	\begin{minipage}[c]{0.4\textwidth}
	\centering
	${}^2$Department of Electrical Engineering\\
	Linköping University
\end{minipage}
	\\[1mm]
	{\tt\small marcus.valtonen\_ornhag@math.lth.se}
\hackspace{-5mm}
}

\maketitle

\begin{abstract}
In this paper, we argue that modern pre-integration methods for inertial
measurement units (IMUs) are accurate enough to ignore the drift
for short time intervals. This allows us to consider a simplified camera model,
which in turn admits further intrinsic calibration. We develop the first-ever
solver to jointly solve the relative pose problem with unknown and equal focal length and radial
distortion profile while utilizing the IMU data. Furthermore, we show significant speed-up compared
to state-of-the-art algorithms, with small or negligible loss in accuracy for partially calibrated setups.

The proposed algorithms are tested on both synthetic and real data, where the latter is
focused on navigation using unmanned aerial vehicles~(UAVs).
We evaluate the proposed solvers on different commercially available low-cost UAVs, and
demonstrate that the novel assumption on IMU drift is feasible in real-life applications.
The extended intrinsic auto-calibration enables us to use distorted input images, making
tedious calibration processes obsolete, compared to current state-of-the-art
methods.\footnote{This work was supported by the strategic research projects ELLIIT and eSSENCE,
the Swedish Foundation for Strategic Research project, Semantic Mapping and Visual Navigation for
Smart Robots (grant no. RIT15-0038), and Wallenberg AI, Autonomous Systems and Software
Program (WASP) funded by Knut and Alice Wallenberg Foundation.
Code available at: \url{https://github.com/marcusvaltonen/DronePoseLib}.}
\hackspace{-5mm}
\end{abstract}


\begin{figure}[t]
\centering
\includegraphics[width=\linewidth]{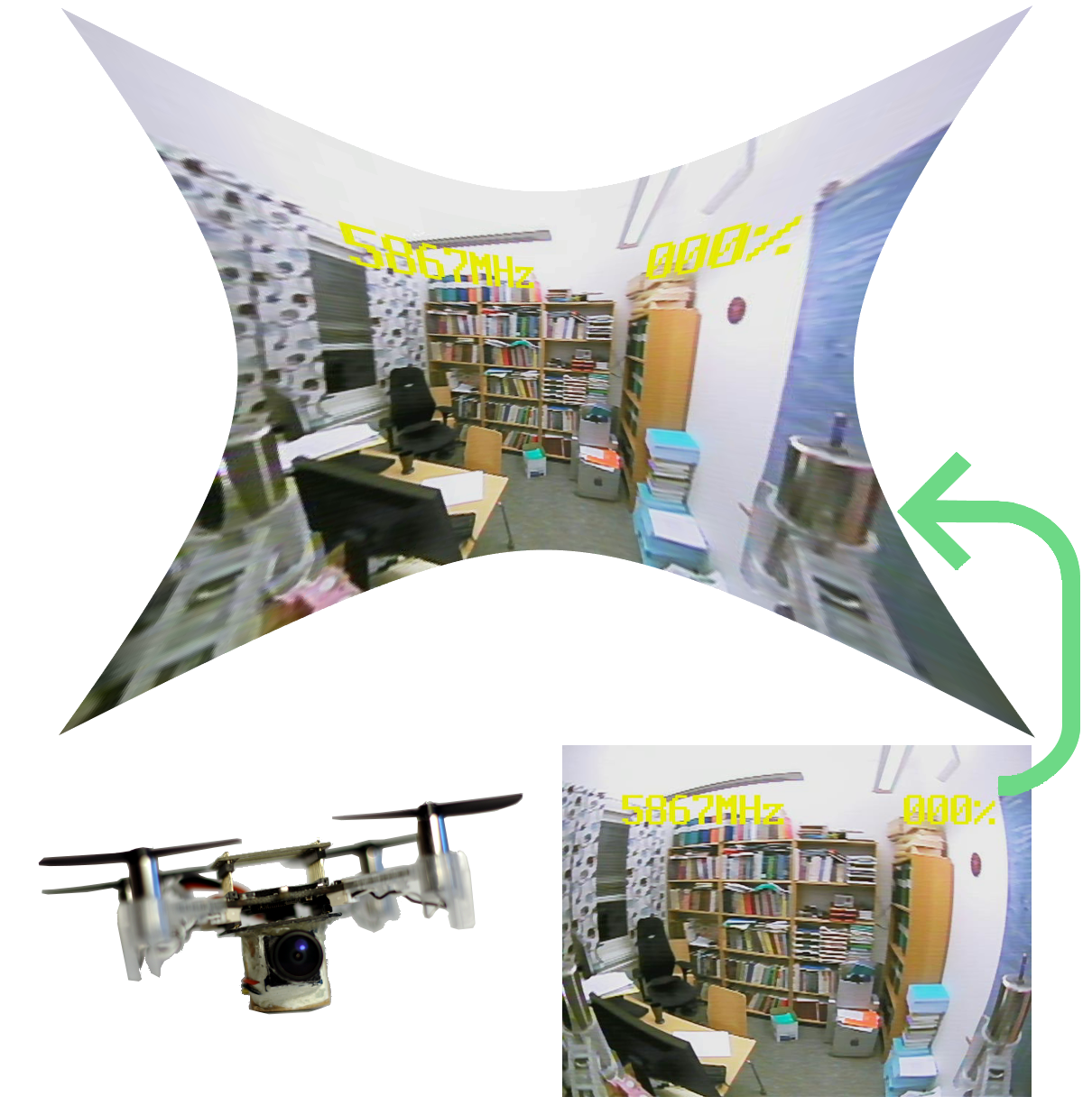}
\caption{The proposed 4-point solver is able to accurately perform radial distortion auto-calibration
for focal length and motion parameters. This is a novel case in the literature and is primarily made feasible
by a clever assumption---to use the complete rotation estimate from pre-integrated IMU data. We
perform experiments with two different UAVs in various difficult scenarios, demonstrating
that this assumption comes with a vast array of
benefits, \eg{}~simpler equations, faster solvers, and little to no loss in accuracy compared to
other state-of-the-art methods.}
\hackspace{-5mm}
\label{fig:front}
\end{figure}
\section{Introduction}
A popular approach in Simultaneous Localization and Mapping (SLAM) is to fuse various
sensor data to increase the performance of the system. A common pair of sensors to combine is
a camera and an IMU. Systems of this kind are labeled as
visual-inertial odometry, and this specific sensor combination is often found on consumer devices, such as
smartphones and UAVs.
As is well-known, the projective relationship between two cameras manifests
itself in the fundamental matrix, independent of the scene geometry. When auxiliary data are known
(\eg{}~IMU data) the number of degrees of freedom decreases and the corresponding fundamental matrix
is constrained, which enables one to
compute it with fewer point correspondences. This potentially reduces the impact of noise;
however, such algebraic constraints can be highly non-trivial to incorporate in a solver. Popular
methods to handle the resulting polynomial systems of equations use theory from algebraic
geometry, \eg{}~the action matrix method~\cite{cox2}, turning the problem
into a generalized eigenvalue problem (GEP)~\cite{kukelova2008bmvc,kukelova-etal-tpami-2012},
and resultant based methods~\cite{bhayani-etal-cvpr-2020}. Regardless
of which method is used, there is still much work in terms of parameterizing the original
problem, as different approaches may yield completely different
results~\cite{larsson-etal-cvpr-2017}.
For the resulting solver to be of any practical use, it must also be numerically stable,
which adds further considerations to the design.

Since modern image sensors often have square-shaped pixels and the lens is sufficiently aligned
such that the
principal point coincides with the optical center, a feasible assumption is to use
partially calibrated cameras, where the only unknown intrinsic parameter is the focal length.
This specific assumption has proven useful in several different real
scenarios including relative pose estimation~\cite{jiang-etal-accv-2014,kuang-etal-cvpr-2014,kukelova-etal-2017-cvpr,ding-etal-cvpr-2020,guan-etal-cvpr-2020}
and absolute pose estimation~\cite{wu-cvpr-2015,larsson-etal-cvpr-2018}.

Although it often comes at the cost of introducing distortion,
having a wide field of view is desirable in many applications.
When working with visual odometry, it is, therefore, a standard
procedure to correct for these undesirable distortion artifacts, which often requires a specific
calibration setup, typically involving a checkerboard pattern. By incorporating a distortion model, as
well as focal length, together with the motion model, one may omit such procedures altogether;
however, due to the difficulty of the problem, no fast and robust minimal solver has yet been proposed.
The main contributions of this paper are:
\begin{itemize}[itemsep=0mm]
\item We take advantage of IMU data to estimate
the full 3D orientation. Under the assumption that the IMU drift is negligible for short time
intervals, the resulting polynomial systems of equations are significantly easier to solve.
\item By using this approach, we are able to treat the partially calibrated case with unknown radial distortion profile
while incorporating the IMU data, resulting in a fast and reliable solver. This is the only
solver to handle this case to date.
\item Furthermore, we show a considerable speed-up compared to other state-of-the-art methods, with
small or insignificant loss in accuracy, when exploiting the assumption of negligible IMU drift. This
benefits low-cost and embedded devices, which constitute the majority of consumer devices where these
algorithms are used in practice.
\end{itemize}

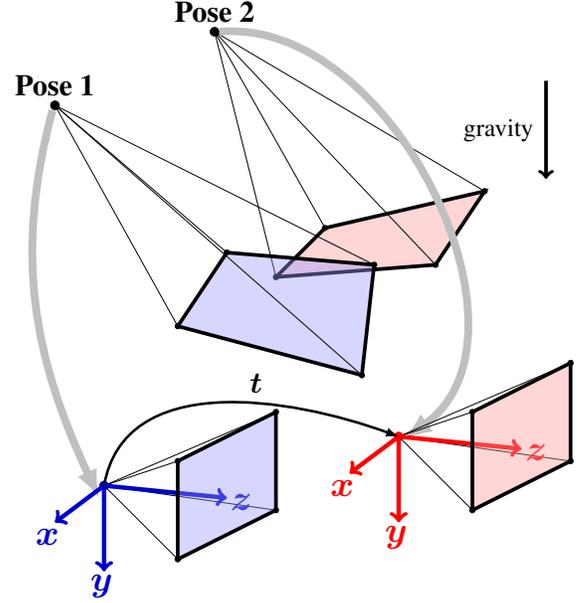
\begin{figure}[t]
\centering
\begin{tikzpicture}
  \filldraw (0,1.5) coordinate (b2a) circle (1.5pt);
  \filldraw (1.0,2.5) coordinate (b2b) circle (1.5pt);
  \filldraw (4.25,3.25) coordinate (b2c) circle (1.5pt);
  \filldraw (3.25,1.75) coordinate (b2d) circle (1.5pt);
  \filldraw (-1.25, 6.5) coordinate (a2) circle (2.5pt) node[yshift=4mm] {\LARGE \textbf{Pose 2}};

  \filldraw [line width = \linethicknessframe, fill=red!40, fill opacity=0.4] (b2a) -- (b2b) -- (b2c) -- (b2d) -- (b2a);
  \draw [line width=\linethickness] (a2) -- (b2a);
  \draw [line width=\linethickness] (a2) -- (b2b);
  \draw [line width=\linethickness] (a2) -- (b2c);
  \draw [line width=\linethickness] (a2) -- (b2d);

  \filldraw (-2,0.5) coordinate (b3a) circle (1.5pt);
  \filldraw (-1,2) coordinate (b3b) circle (1.5pt);
  \filldraw (2,1.75) coordinate (b3c) circle (1.5pt);
  \filldraw (1.75,-0.5) coordinate (b3d) circle (1.5pt);
  \filldraw (-4.5, 5) coordinate (a3) circle (2.5pt) node[yshift=4mm] {\LARGE \textbf{Pose 1}};

  \filldraw [line width = \linethicknessframe, fill=blue!40, fill opacity=0.4] (b3a) -- (b3b) -- (b3c) -- (b3d) -- (b3a);
  \draw [line width=\linethickness] (a3) -- (b3a);
  \draw [line width=\linethickness] (a3) -- (b3b);
  \draw [line width=\linethickness] (a3) -- (b3c);
  \draw [line width=\linethickness] (a3) -- (b3d);

  \draw[->, line width=\linethicknessframe] (5.5,5.5) -- (5.5,3.5) node [midway, left,xshift=-1mm] {\Large gravity};
  \filldraw[white] (-6.7,-5) coordinate (dummy) circle (0pt);  

  \coordinate[yshift=-1mm,xshift=-0.5mm] (specRoot) at (a3);
  \coordinate (testTreeRoot) at (-3.6,-3);
  \draw[-latex, gray!50!white, line width=1ex]  (specRoot) to[in=115,out=255] (testTreeRoot);

  \coordinate[yshift=0mm,xshift=1mm] (testTreeRoot2) at (a2);
  \coordinate (specRoot2) at (2.6,-1.8);
  \draw[latex-, gray!50!white, line width=1ex]  (specRoot2) to[in=5,out=35] (testTreeRoot2);

  \begin{scope}[yshift=2cm]
  \filldraw[blue!80!black] (-3.5, -4.75  ) coordinate (d) circle (2.5pt);
  \filldraw (-2,-6.25) coordinate (da) circle (1.5pt);
  \filldraw (-2,-4.25) coordinate (db) circle (1.5pt);
  \filldraw (0,-5.25) coordinate (dd) circle (1.5pt);
  \filldraw (0,-3.25) coordinate (dc) circle (1.5pt);
  \draw [line width=\linethickness] (d) -- (da);
  \draw [line width=\linethickness] (d) -- (db);
  \draw [line width=\linethickness] (d) -- (dc);
  \draw [line width=\linethickness] (d) -- (dd);
  \draw[->,line width=0.9mm, blue!80!black] (d) -- (-3.5,-6.5) node[xshift=-0.5mm, yshift=-3mm] {\huge $\bm{y}$};
  \draw[->,line width=0.9mm, blue!80!black] (d) -- (-1,-5) node[xshift=3mm, yshift=-0.8mm] {\huge $\bm{z}$};
  \draw[->,line width=0.9mm, blue!80!black] (d) -- (-4.5,-5.5) node[xshift=-1.5mm, yshift=-3mm] {\huge $\bm{x}$};
  \filldraw [line width = \linethicknessframe, fill=blue!40, fill opacity=0.4] (da) -- (db) -- (dc) -- (dd) -- (da);
  \end{scope}

  \begin{scope}[xshift=6cm,yshift=3cm]
  \filldraw[red] (-3.5, -4.75  ) coordinate (e) circle (2.5pt);
  \filldraw (-2,-6.25) coordinate (ea) circle (1.5pt);
  \filldraw (-2,-4.25) coordinate (eb) circle (1.5pt);
  \filldraw (0,-5.25) coordinate (ed) circle (1.5pt);
  \filldraw (0,-3.25) coordinate (ec) circle (1.5pt);
  \draw [line width=\linethickness] (e) -- (ea);
  \draw [line width=\linethickness] (e) -- (eb);
  \draw [line width=\linethickness] (e) -- (ec);
  \draw [line width=\linethickness] (e) -- (ed);
  \draw[->,line width=0.9mm, red] (e) -- (-3.5,-6.5) node[xshift=-0.5mm, yshift=-3mm] {\huge $\bm{y}$};
  \draw[->,line width=0.9mm, red] (e) -- (-1,-5) node[xshift=3mm, yshift=-0.8mm] {\huge $\bm{z}$};
  \draw[->,line width=0.9mm, red] (e) -- (-4.5,-5.5) node[xshift=-1.5mm, yshift=-3mm] {\huge $\bm{x}$};
  \filldraw [line width = \linethicknessframe, fill=red!40, fill opacity=0.4] (ea) -- (eb) -- (ec) -- (ed) -- (ea);
  \end{scope}

  \draw[latex-, black, line width=1.5pt]  (e) to[in=80,out=160] (d) node[midway, above, yshift = -1.0cm,xshift=-4mm] {\LARGE$\mat{t}$};
\end{tikzpicture}
\caption{Assume the IMU measurements are accurate,~\ie{}~the accelerometer and gyroscope data
can be used to accurately estimate the relative orientation between two consecutive views.
Then the only unknown extrinsic parameter is the translation vector between the poses.}
\label{fig:problem_geometry}
\end{figure}

\section{Previous Work}
\subsection{Visual-Inertial Odometry}
The calibrated visual-inertial problem of relative pose is well-studied~\cite{li-etal-iros-2013,fraundorfer-etal-eccv-2010,sweeney-etal-3dv-2014,naroditsky-etal-tpami-2012} and efficient solvers exist.
If we assume that the gravity direction is aligned with the \mbox{$y$-axis}, the corresponding essential matrix
(after alignment) is given by $\Ey \sim \skewmat{\mat{t}} \mat{R}_y$,
or explicitly,
\begin{equation}
    \Ey =
    \left[\begin{smallmatrix}
                        -t_y\sin{\phi} & -t_z &                 t_y\cos{\phi} \\
         t_z\cos{\phi} + t_x\sin{\phi} &   0  & t_z\sin{\phi} - t_x\cos{\phi} \\
                        -t_y\cos{\phi} &  t_x &                -t_y\sin{\phi}
    \end{smallmatrix}\right],
\end{equation}
where
\begin{equation}
    \Ry =
    \begin{bmatrix}
        \cos{\phi} & 0 & \sin{\phi} \\
                 0 & 1 &          0 \\
       -\sin{\phi} & 0 & \cos{\phi}
    \end{bmatrix},
\end{equation}
and~$\mat{t}=(t_x,\,t_y,\,t_z)$.
This makes it possible to use a parameterization with six elements,
\begin{equation}\label{eq:Eyparams}
    \Ey =
    \begin{bmatrix}
        e_1 & e_2 & e_3 \\
        e_4 &   0 & e_5 \\
       -e_3 & e_6 & e_1
    \end{bmatrix}\;.
\end{equation}
Since we have four degrees of freedom (three translation elements and one
angle), the elements~$e_i$ of~\eqref{eq:Eyparams} are not independent. In fact,
one can check that they must obey the (modified) Demazure
equations, also known as the trace
constraint,~\mbox{$2\Ey\Ey^\T\Ey - \tr(\Ey\Ey^\T)\Ey=0$}, or explicitly,
\begin{equation}\label{eq:demazure}
\begin{aligned}
    e_2^2 - e_4^2 - e_5^2 + e_6^2 &= 0, \\
    e_1e_2e_6 + e_1e_4e_5 + e_3e_5^2 - e_3e_6^2 &= 0, \\
    e_1e_4^2 - e_1e_6^2 - e_2e_3e_6 + e_3e_4e_5 &= 0,
\end{aligned}
\end{equation}
as well as the rank constraint~$\det(\Ey)=0$,
\begin{equation}\label{eq:rank}
    e_1e_2e_4 + e_1e_5e_6 + e_2e_3e_5 - e_3e_4e_6 = 0\;.
\end{equation}
These constraints were used in~\cite{fraundorfer-etal-eccv-2010} to build a minimal
solver for the calibrated case\footnote{In~\cite{fraundorfer-etal-eccv-2010} they align the $z$-axis
with the gravity instead.}.
We also note an easy decomposition into rotation and translation components, given
by~$e_1^2+e_3^2=t_y^2$.
In~\cite{guan-etal-cvpr-2020} it was shown that the minimal case can be solved using a single
affine correspondence.

The problem becomes more difficult when adding an unknown focal length. Without
any IMU data available, but still considering the partially calibrated case with only
unknown focal length, the corresponding fundamental matrix has six degrees of freedom. This
problem, therefore, requires a minimal case of six point correspondences, with the current
state-of-the-art solver by Kukelova~\etal{}~\cite{kukelova-etal-2017-cvpr}.
Ding~\etal{}~\cite{ding-etal-cvpr-2020} proposed a minimal solver for two
partially calibrated cases while incorporating the IMU data. This was done by explicitly
parameterizing the rotation about the gravity direction, and turning the problem into a generalized
eigenvalue problem (GEP).

When assuming $\mat{R}_y=\mat{I}$, see~\cref{fig:problem_geometry}, the essential matrix is~$\mat{E}=[\mat{t}]_\times$, which
makes the governing equations significantly easier. The minimal calibrated case requires only two point correspondences, and the epipolar constraint for a single pair of
point correspondences $\mat{x}\leftrightarrow\mat{x}'$, is given by
\begin{equation}\label{eq:lineart}
    \mat{x}'^\T\mat{E}\mat{x} = 0
    \Leftrightarrow
    (\mat{x}\times\mat{x}')^\T\mat{t} = 0\;.
\end{equation}
In~\cite{fredriksson-etal-cvpr-2016} it was also shown that the non-minimal case can be solved
with global optimality guarantees.

\subsection{Relative Pose with Unknown Distortion Profile}

When constructing minimal solvers, it is often desirable to use as few parameters as possible.
This increases robustness in RANSAC-like frameworks, as fewer
iterations are needed in order to select a sample free from outliers.
The one-parameter division model~\cite{fitzgibbon2001}, has
therefore been frequently used, as it performs well with only a single parameter for a large
variety of different lenses.
In this model, the radially distorted image point~$\mat{x}=(x,\,y,\,1)$ is assumed to be mapped
to its corrected counterpart~$\hat{\mat{x}}$ through the following parametric relation
\begin{equation}\label{eq:divisionmodel}
    \hat{\mat{x}} = f(\mat{x},\,\lambda) = \begin{bmatrix}
        x \\ y \\ 1 + \lambda(x^2 + y^2)
    \end{bmatrix},
\end{equation}
where $\lambda$ controls the level of distortion. It has been used successfully in a number
of applications~\cite{kukelova2015,pritts2017,kukelova-etal-2017-cvpr,pritts2018,larsson-etal-cvpr-2018,valtonen-ornhag-etal-wacv-2021}.

The case of relative pose with unknown focal length and unknown distortion parameter is known
to be hard. The two-sided problem, \ie{}~equal and unknown focal length and radial distortion
parameter, was first studied in~\cite{jiang-etal-accv-2014}; however, by today's standards, one cannot say that the proposed solver has much practical use: the elimination template size
is very large,~$886\times 1011$, with 68 putative solutions, and a reported runtime of 400 ms.
In~\cite{larsson-etal-cvpr-2017} the elimination template size was reduced to $581\times 862$
using their proposed reduction step; however, no analysis of the numerical stability was
performed.
Regardless, it remains impractical for real-life applications, as the size is still exceedingly
large. There has been some theoretical work on the problem, and more generally on distortion
varieties~\cite{kileel-etal-2018}; however, no viable real-time solver for the case exists.

The one-sided case, \ie{}~with one calibrated camera and one camera with unknown focal length and
radial distortion parameter, has been studied further. The first solver was
introduced in~\cite{kuang-etal-cvpr-2014}, but was not numerically stable and the elimination template size
was quite large,~$200\times 231$. It has later been improved in~\cite{kukelova-etal-2017-cvpr},
and is now both numerically stable and fast, with an elimination template size of~$51\times 70$.
The one-sided case, however, is mostly artificial, as it assumes one of the cameras to be calibrated,
which limits the applicability of the method severely.

To the best of our knowledge, the relative pose problem with unknown and equal focal length and
radial distortion parameter incorporating IMU data has not been solved. We will solve
this case and show that the resulting solver is extremely fast compared to the methods discussed
in this section, with an elimination template size of merely~$10\times 21$. This
is done using a special assumption, which we shall discuss next.

\section{Why Ignore the IMU Drift?}\label{sec:whyignore}

When measurements from the accelerometer and gyroscope are combined in an orientation filter \cite{ekf_orientation, ekf_orientation, ukf_orientation, magdwick, complimentary_filter} the gravity direction is preserved; however, the yaw angle begins to drift.
Because of this, most visual-inertial models try to incorporate an unknown angle about the gravity
direction~\cite{li-etal-iros-2013,fraundorfer-etal-eccv-2010,sweeney-etal-3dv-2014,naroditsky-etal-tpami-2012,ding-etal-2019-iccv,ding-etal-cvpr-2020,guan-etal-cvpr-2020,valtonen-ornhag-etal-icpr-2021}.
Already in the calibrated case the governing equations~\eqref{eq:demazure}--\eqref{eq:rank}
are non-trivial and quadratic or cubic in nature. Parameterizing the rotation matrix will also result in at least second order equations.

\begin{figure*}[t]
\centering
\begin{tikzpicture}[>=latex]
    \begin{axis}[
        width=0.4765\textwidth,
        height=0.25\textwidth,
        axis x line*=bottom,
        axis y line*=left,
        ylabel={Frequency} ,
        ymin = 0,
        ymax = 1500,
        xmin = -20,
        xmax = 0,
        xtick = {-20, -15, -10, -5, 0},
        xticklabels={$10^{-20}$,$10^{-15}$,$10^{-10}$,$10^{-5}$,$1$},
        xlabel={Fundamental matrix error},
  ]
        \addplot[very thick,color=ourcolor] table {data/error_fEf_no_rot_3pt_fundamental.dat};
        \addplot[very thick,color=ourcolor2] table {data/error_frEfr_no_rot_4pt_fundamental.dat};
        \addplot[very thick,densely dotted, color=dingcolor] table {data/error_fEf_ding_4pt_fundamental.dat};
        \addplot[very thick,dashed, color=kukelovacolor] table {data/error_fEf_kukelova_6pt_fundamental.dat};
        \addplot[very thick,dashed, color=jiangcolor] table {data/error_jiang_frEfr_7pt_fundamental.dat};
    \end{axis}
\end{tikzpicture}
\begin{tikzpicture}[>=latex]
    \begin{axis}[
        width=0.4765\textwidth,
        height=0.25\textwidth,
        axis x line*=bottom,
        axis y line*=left,
        ylabel={Frequency} ,
        ymin = 0,
        ymax = 1500,
        xmin = -20,
        xmax = 0,
        xtick = {-20, -15, -10, -5, 0},
        xticklabels={$10^{-20}$,$10^{-15}$,$10^{-10}$,$10^{-5}$,$1$},
        xlabel={Focal length error},
  ]
        \addplot[very thick,color=ourcolor] table {data/error_fEf_no_rot_3pt_focal.dat};
        \addplot[very thick,color=ourcolor2] table {data/error_frEfr_no_rot_4pt_focal.dat};
        \addplot[very thick,densely dotted, color=dingcolor] table {data/error_fEf_ding_4pt_focal.dat};
        \addplot[very thick,dashed, color=kukelovacolor] table {data/error_fEf_kukelova_6pt_focal.dat};
        \addplot[very thick,dashed, color=jiangcolor] table {data/error_jiang_frEfr_7pt_focal.dat};
    \end{axis}
\end{tikzpicture}
\includestandalone[width=0.8\linewidth]{graphs/synth/histogram_legend}
\caption{Error histogram for 10,000 randomly generated problem instances. The proposed 4-point
method and the 7-point method~\cite{jiang-etal-accv-2014} also solve for an unknown radial distortion
parameter.}
\label{fig:compare_stability}
\end{figure*}
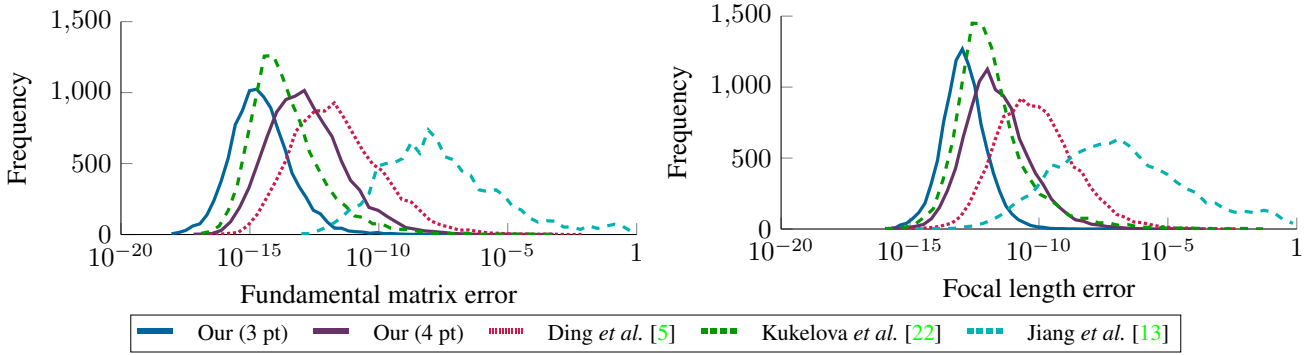

Instead of filtering, another approach is to simply integrate the gyroscopic data to obtain the relative orientation~\cite{pre_integration}. This eliminates error sources that potentially are introduced while fusing the gyroscopic data with the accelerometer data, \eg{}~Coriolis forces---when the IMU is not in the center of rotation---or acceleration due to motion. Both corrupt the measurement of the gravity direction. Over time integration errors, sensor bias and sensor noise will lead to drift in all axes; however, for short time intervals, this drift is very small. Additionally, if the IMU sensor has been stationary at some point in time---which is a reasonable assumption for UAVs that start from a ground position---the gyroscopic bias can be observed. The bias changes very slowly, even in low-cost consumer-available IMUs, and the observed bias can therefore be used to compensate for it in the remaining part of the sequence. In summary, we obtain very accurate relative rotations by simply integrating the gyroscopic data between camera pairs.

Valtonen Örnhag~\etal{}~\cite{valtonen-ornhag-etal-wacv-2021} proposed using orientation filters to estimate the camera rotation and thereby benefit from a relaxed problem. They showed that this allowed them to perform radial distortion correction, while still maintaining speed and accuracy.
The method, however, relied on a homography-based approach, requiring planar objects in the scene geometry, thus limiting the applicability of the method.

Temporarily losing the gravity direction is not a major concern for the relative pose problem.
In a complete SLAM framework, one would typically perform  a visual-inertial initialization step \cite{Martinelli2014, VI_step1, 7817784, 9197334} which recovers the gravity direction as well as metric scale and bias. By trusting the IMU data we note the following:
\begin{enumerate}[label={(\roman*)}]
\item we remove one degree of freedom from the camera parameterization,
\item the relative pose problem becomes linear in the
unknown translation, according to~\eqref{eq:lineart},
\item we open up the possibility for further intrinsic calibration and still perform fast
and accurate in real-time applications.
\end{enumerate}
In the next section, we will show how this is done in practice.

\def\boxplotheight{8.0cm}
\def\boxplotwidth{7.5cm}
\begin{figure*}[t]
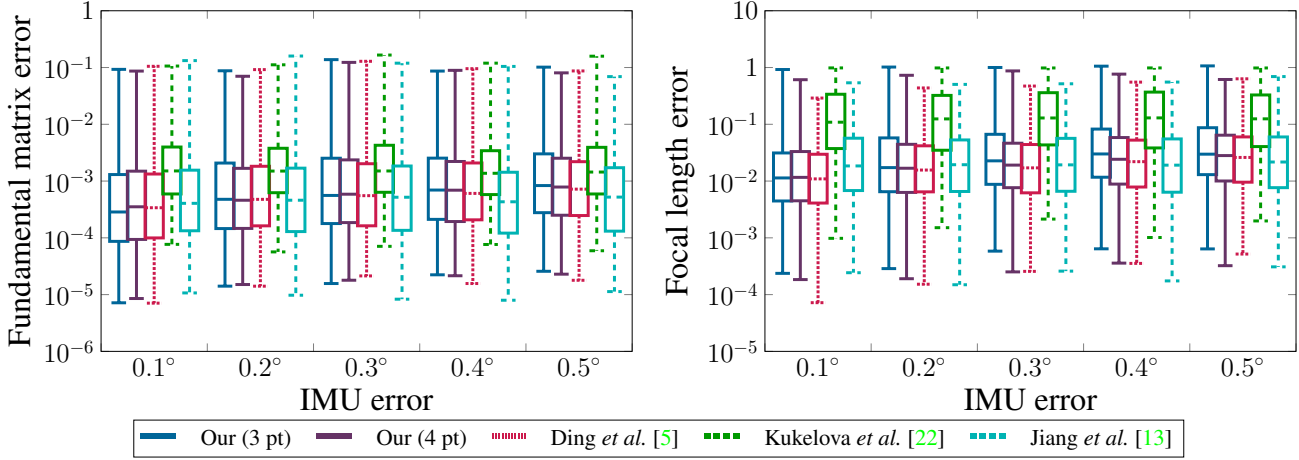

\centering
\begin{tikzpicture}
\begin{axis}[
boxplot/draw direction=y,
ylabel={Fundamental matrix error},
xlabel={IMU error},
label style={font=\LARGE},
height=\boxplotheight,
width=\boxplotwidth,
xmin=0,xmax=5,
ymin=-6,ymax=0,
cycle list={
    {ourcolor, \thicknesslineplot},
    {ourcolor2, \thicknesslineplot},
    {dingcolor, densely dotted, \thicknesslineplot},
    {kukelovacolor, dashed, \thicknesslineplot},
    {jiangcolor, dashed, \thicknesslineplot}
},
boxplot/every box/.style={solid},
boxplot prepared={
        draw position={1/6 + floor(\plotnumofactualtype/5) + 1/6*fpumod(\plotnumofactualtype,5)},
        box extend=0.172,
},
x=2cm,
xtick={0,1,...,5},
xticklabels={%
        {$0.1^\circ$},%
        {$0.2^\circ$},%
        {$0.3^\circ$},%
        {$0.4^\circ$},%
        {$0.5^\circ$},%
},
x tick label as interval,
x tick label style={
        text width=2.5cm,
        align=center,
        font=\Large
},
ytick={-6,-5,...,0},
yticklabels={%
        {$10^{-6}$},%
        {$10^{-5}$},%
        {$10^{-4}$},%
        {$10^{-3}$},%
        {$10^{-2}$},%
        {$10^{-1}$},%
        {$1$},%
},
y tick label style={
        font=\Large,
},
]
\input{./graphs/data/noise_case_realistic0_fundamental.dat}
\end{axis}
\begin{axis}[
xshift=12.5cm,
boxplot/draw direction=y,
ylabel={Focal length error},
xlabel={IMU error},
label style={font=\LARGE},
height=\boxplotheight,
width=\boxplotwidth,
xmin=0,xmax=5,
ymin=-5,ymax=1,
cycle list={
    {ourcolor, \thicknesslineplot},
    {ourcolor2, \thicknesslineplot},
    {dingcolor, densely dotted, \thicknesslineplot},
    {kukelovacolor, dashed, \thicknesslineplot},
    {jiangcolor, dashed, \thicknesslineplot}
},
boxplot/every box/.style={solid},
boxplot prepared={
        draw position={1/6 + floor(\plotnumofactualtype/5) + 1/6*fpumod(\plotnumofactualtype,5)},
        box extend=0.172,
},
x=2cm,
xtick={0,1,...,5},
xticklabels={%
        {$0.1^\circ$},%
        {$0.2^\circ$},%
        {$0.3^\circ$},%
        {$0.4^\circ$},%
        {$0.5^\circ$},%
},
x tick label as interval,
x tick label style={
        text width=2.5cm,
        align=center,
        font=\Large
},
ytick={-5,...,1},
yticklabels={%
        {$10^{-5}$},%
        {$10^{-4}$},%
        {$10^{-3}$},%
        {$10^{-2}$},%
        {$10^{-1}$},%
        {$1$},%
        {$10$},%
},
y tick label style={
        font=\Large,
},
]
\input{./graphs/data/noise_case_realistic0_focal.dat}
\end{axis}


\end{tikzpicture}
\includestandalone[width=0.8\linewidth]{graphs/boxplot/box_legend}
\caption{Error for various IMU noise levels. A total of 1,000 randomly generated problem instances are used per noise level.}
\label{fig:noise_exp}
\end{figure*}

\section{Consequences of Ignoring the IMU Drift}
We construct two solvers based on our simplifying assumption. Note that the
derivations are quite short, which is mainly due to the linear dependence on the translation vector.

\subsection{Unknown and Equal Focal Length (3-point)}
For the case of unknown and equal focal length, the fundamental matrix is given
by~$\mat{F} = \mat{K}^{-1}[\mat{t}]_\times\mat{R}\mat{K}^{-1}$, where
$\mat{R}=\mat{R}_{\text{imu}}^{(2)}\mat{R}_{\text{imu}}^{(1)\T}$ is the relative orientation.
By parameterizing~$\mat{K}^{-1}=\diag(1,\,1,\,w)$, we have four unknowns---the translation~$\mat{t}$
and~$w$. The epipolar constraints $\mat{x}_i^{'\T}\mat{Fx}_i=0$, for~$i=1,2,3$, yield three
equations, which are linear in~$\mat{t}$. Therefore, the resulting system can be written
as
\begin{equation}\label{eq:Mt_3pt}
    \mat{M}(w)\mat{t} = \mat{0},
\end{equation}
where~$\mat{M}(w)$ depends only on~$w$. Even with non-degenerate configurations, the matrix
$\mat{M}\in\R^{3\times 3}$ cannot have full rank, hence~$\det(\mat{M})=0$. This yields
a single quartic equation in the unknown~$w$, which can be solved efficiently using
the quartic root finding formula. Consequently, we have four putative solutions, from which the
translation~$\mat{t}$ can be retrieved by finding the null space of the $3\times 3$ matrix
$\mat{M}(w^*)$, where~$w^*$ is a putative solution. For the $3\times 3$ case one can do this
without resulting to costly SVD computations~\cite{McAdams2011ComputingTS,Gast2016ImplicitshiftedSQ}.

\subsection{Unknown and Equal Focal Length and Radial Distortion Parameter (4-point)}
Assuming the one-dimensional division model~\eqref{eq:divisionmodel}, the (modified)
epipolar constraint is given by
\begin{equation}\label{eq:modepi}
    f(\mat{x}'_i,\,\lambda)^\T\mat{K}^{-1}[\mat{t}]_\times\mat{R}\mat{K}^{-1} f(\mat{x}_i,\,\lambda) = 0,
\end{equation}
for a pair of corresponding~$\mat{x}_i\leftrightarrow\mat{x}'_i$,
where~$\mat{K}^{-1} = \diag(1,\,1,\,w)$, with $w\neq 0$, and $\lambda$ is the unknown distortion
parameter.
As in the previous case, we may utilize the fact that~\eqref{eq:modepi} is linear in~$\mat{t}$,
hence
\begin{equation}
    \mat{M}(w,\,\lambda)\mat{t} = \mat{0},
\end{equation}
where~$\mat{M}\in\R^{4\times 3}$. This can be viewed as seeking the non-trivial nullspace of~$\mat{M}$, which, in turn, implies that all $3\times 3$ subdeterminants of~$\mat{M}$ must vanish. Since
there are four such subdeterminants, we can reduce the problem to four polynomial equations in two
unknowns, $w$ and~$\lambda$. Furthermore, we need to exclude non-physical solutions corresponding
to~$w=0$, as it turns out that there are infinitely many solutions if we allow these.
This can be accomplished by saturating the corresponding ideal and can be
done using the automatic Gröbner basis generator proposed in~\cite{larsson2017ICCV}.
It turns out that the remaining system
has eleven solutions in general; however, in practice, the most common case is that 4--6 solutions
are real-valued.
By using the action matrix method~\cite{cox2}, we are able to construct an elimination template
of size~$10\times 21$, by using the basis heuristic proposed in~\cite{larsson2018cvpr}.

\section{Time Complexity}
To compare timings in a fair and accurate way all solvers are implemented in C++ using the
Eigen~\cite{eigen} library, with the same compilation flags and setup.
The only exception is that the original solver by~Jiang~\etal{}~\cite{jiang-etal-accv-2014}
is in native \mbox{MATLAB}, hence significantly slower. To make a fair comparison we use the
slightly faster (but numerically unstable) solver proposed in~\cite{larsson-etal-cvpr-2017} which is
available in C++. Note, therefore, that the real execution time for the original method
would be even larger than reported. The faster solver is only used for timing,
and the original solver is used for all other experiments.

To simulate a realistic
environment we use a Raspberry Pi~4 to measure the execution time, as it is a fair approximation
of hardware you can expect on an embedded device running these algorithms.
The results are shown in~\cref{tab:exectime}.

\begin{table}[h!]
\begin{center}
\caption{Mean execution time on a Raspberry Pi~4 for 10,000 randomly generated problem instances in C++. We also show the number of solutions for each problem. These will have to be evaluated (or
at least a subset) in a
RANSAC-like system, hence affect the total execution time.}
\label{tab:exectime}
\vspace{1mm}  
\begin{tabular}{lll}
\hline
Author                                          & Time ($\mu$s) & No. Solutions\\ \hline
Our~(3-point)                                       & 6   & 4  \\
Ding~\etal{}~\cite{ding-etal-cvpr-2020}         & $4815^\dagger$ & 20 \\
Kukelova~\etal{}~\cite{kukelova-etal-2017-cvpr} & 363               & 15  \\ \hline
Our~(4-point)                                     & 1290 & 11 \\
Jiang~\etal{}~\cite{jiang-etal-accv-2014}       & $1,260,700^\ddagger$ & 68 \\ \hline
\end{tabular}
\end{center}
{\footnotesize
$\dagger$: C++ implementation received from the authors of~\cite{ding-etal-cvpr-2020}.\\
$\ddagger$: Based on the smaller template reported in~\cite{larsson-etal-cvpr-2017} which is
numerically unstable. The original solver would be even slower.}
\end{table}

Comparing the proposed 3-point solver to the state-of-the-art solver by
Ding~\etal{}~\cite{ding-etal-cvpr-2020}---essentially solving the same problem, with the exception that we ignore the potential IMU drift---our solver is more than
800$\times$ faster. In addition, the proposed 4-point solver including focal length and radial distortion
correction is a factor $3.7\times$ faster than the solver by Ding~\etal{}~\cite{ding-etal-cvpr-2020},
and significantly faster than the solver
by~\cite{jiang-etal-accv-2014}; in fact, it is roughly~1000$\times$ faster, bridging the gap
from what was considered a theoretically interesting case to something that can be applied
in practice.

Let us emphasize the practical implications of simultaneously estimating the distortion parameter:
the added intrinsic calibration liberates the user from time-consuming calibration procedures.
This allows UAV operators (and those of other visual-inertial systems) to change optics out in the field,
with no intermediate setup procedures or specific requirements needed.

\def\www{0.235\textwidth}
\begin{figure*}[t]
\centering
\begin{tabular}{cccc}
\emph{Basement} &
\emph{Carpet} &
\emph{Indoor} &
\emph{Outdoor} \\
\includegraphics[width=\www]{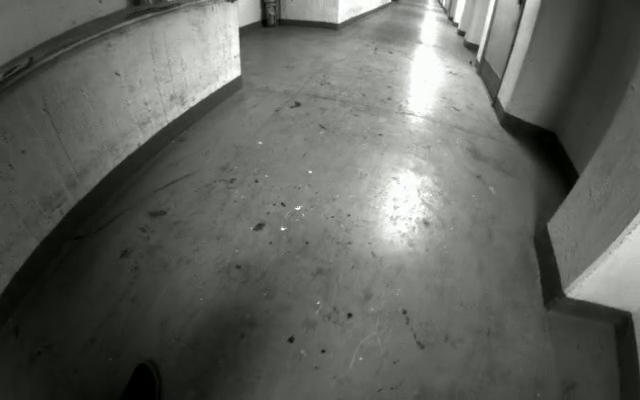} &
\includegraphics[width=\www]{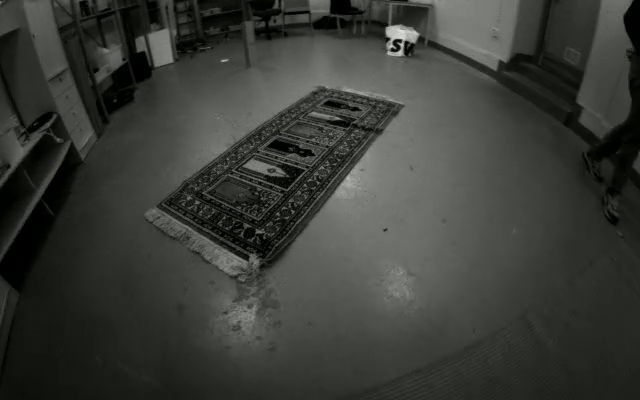} &
\includegraphics[width=\www]{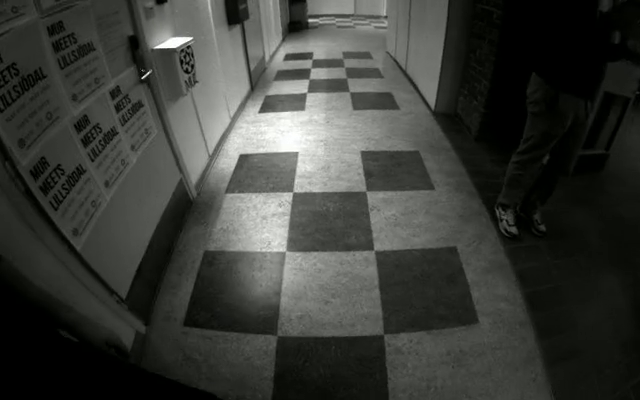} &
\includegraphics[width=\www]{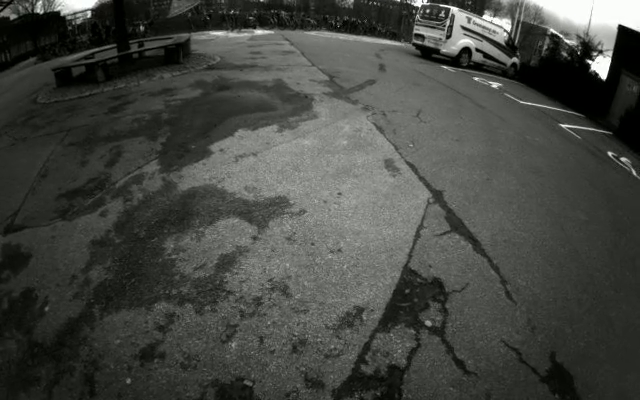} \\
\emph{Bicycle lane} &
\emph{Building} &
\emph{Corridor} &
\emph{Office} \\
\includegraphics[width=\www]{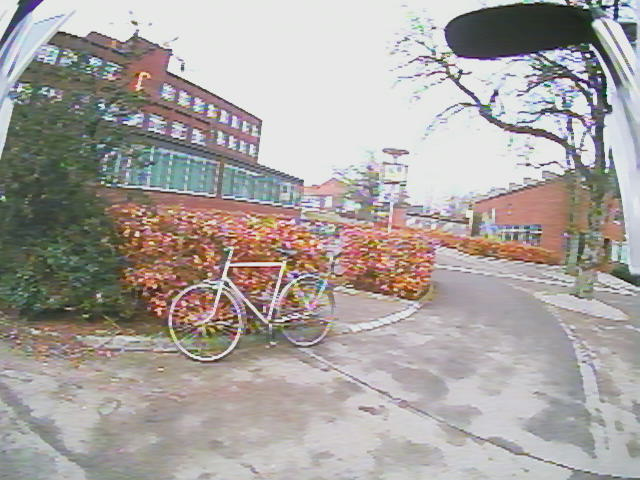} &
\includegraphics[width=\www]{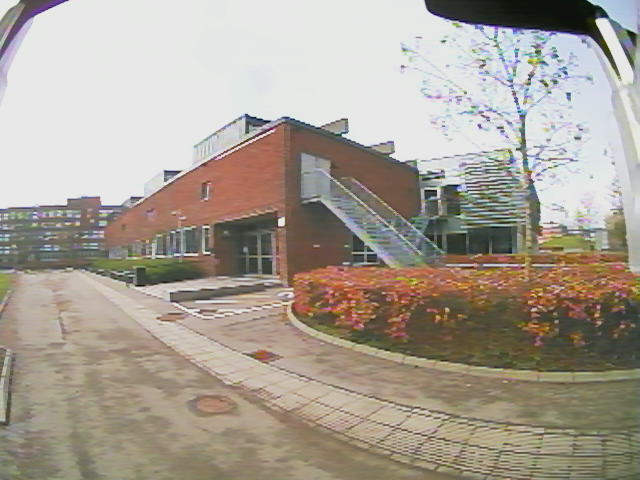} &
\includegraphics[width=\www]{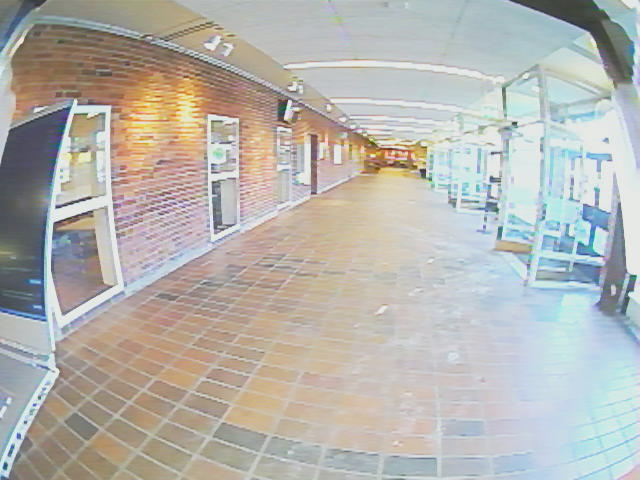} &
\includegraphics[width=\www]{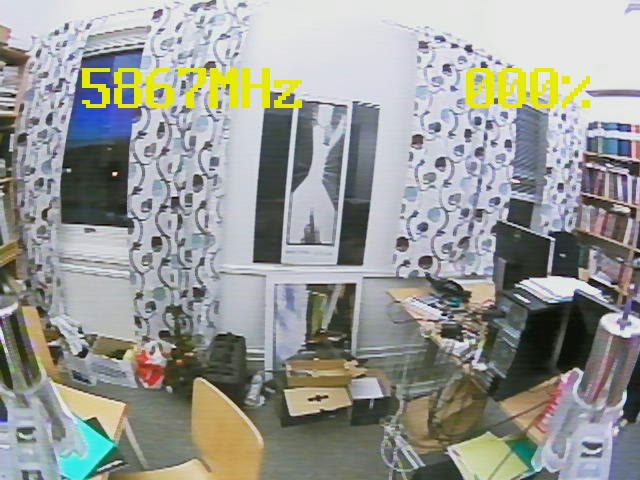}
\end{tabular}
\caption{(\emph{Top row}): Example images from the dataset~\cite{valtonen-ornhag-etal-icpr-2021}.
(\emph{Bottom row}): Images from the new dataset using the \emph{Crazyflie~2.0}. Note that all images suffer from radial distortion to some extent.
}
\hackspace{-1mm}  
\label{fig:thumbnails}
\end{figure*}

\section{Synthetic Experiments}
In this section, we test the numerical stability and noise sensitivity of our proposed methods
compared to the current state-of-the-art.
The competing methods are the 4-point solver by Ding~\etal{}~\cite{ding-etal-cvpr-2020},
the 6-point solver by Kukelova~\etal{}~\cite{kukelova-etal-2017-cvpr}
and the 7-point solver by Jiang~\etal{}~\cite{jiang-etal-accv-2014}.
We found that the solver proposed in~\cite{larsson-etal-cvpr-2017},
which was reported to have a smaller template size, was numerically unstable; hence
we use the original method instead.

In order to get a realistic setup, random synthetic
scene points with a positive depth in front of the
cameras were generated. Specifically,
the scene points $(X,\,Y,\,Z)$ were uniformly distributed with $X,Z\in[-3,\,3]$, the depth $Y\in[3,\,8]$, and
focal length $f\in[300,\, 3000]$. This mimics the setup used in~\cite{ding-etal-cvpr-2020}.
Furthermore, the orientations are random and facing the scene.
The image points are then obtained by projecting the scene points through the cameras, and
the orientations of the cameras are used as input to the visual-inertial solvers.
To increase numerical stability, the image point correspondences were normalized; this
was done in the same way for all solvers.
The error distribution for noise-free data is shown in~\cref{fig:compare_stability}.
Here, all methods perform well, with a slight advantage for our 3-point method. Note that we
include radial distortion for the proposed \mbox{4-point} solver and the 7-point solver in this case as well;
however, the error histograms are similar and therefore omitted.

We proceed by analyzing noise sensitivity in various situations.
We test all methods on synthetic data without radial distortion (including the 4-point and 7-point solvers),
and add a pixel noise relative to focal length (Gaussian noise with zero mean and standard deviation $1080/f$ pixels,
where $f$ is the ground truth focal length).
Furthermore, we add noise
to the IMU measurements---here we add noise on \emph{all} angles, including the yaw angle.
The noise interval is meant to cover the precision of a low-cost IMU, with a maximal error
of approximately $0.5^\circ$, as reported in~\cite{kukelova-etal-accv-2010}.
The results are presented in~\cref{fig:noise_exp}. Note that both our methods perform better than
the competing methods in terms of fundamental matrix recovery for smaller noise levels, and the
state-of-the-art method~\cite{ding-etal-cvpr-2020} only performs slightly better for larger noise
levels. This is primarily since it is capable of correcting for the error about the gravity
direction. We have, however, found empirically on real-data that the lower noise levels are
dominating the input data under certain conditions. This holds true even for low-cost devices,
hence do not pose a practical issue.

\section{Real Data}
To demonstrate the applicability of our assumption, and the solvers based on it,
we have used the challenging dataset~\cite{valtonen-ornhag-etal-icpr-2021} consisting of
various indoor and outdoor scenes with predominantly planar surfaces.
The scenes were captured using a mid-sized UAV (500 g, $170\times 240\times 40$ mm) equipped with
a monochrome global shutter camera (OV9281) recorded with resolution $640\times400$, equipped with an onboard
IMU of model MPU-9250.
\begin{figure*}[t]
\centering
\includegraphics[width=\linewidth]{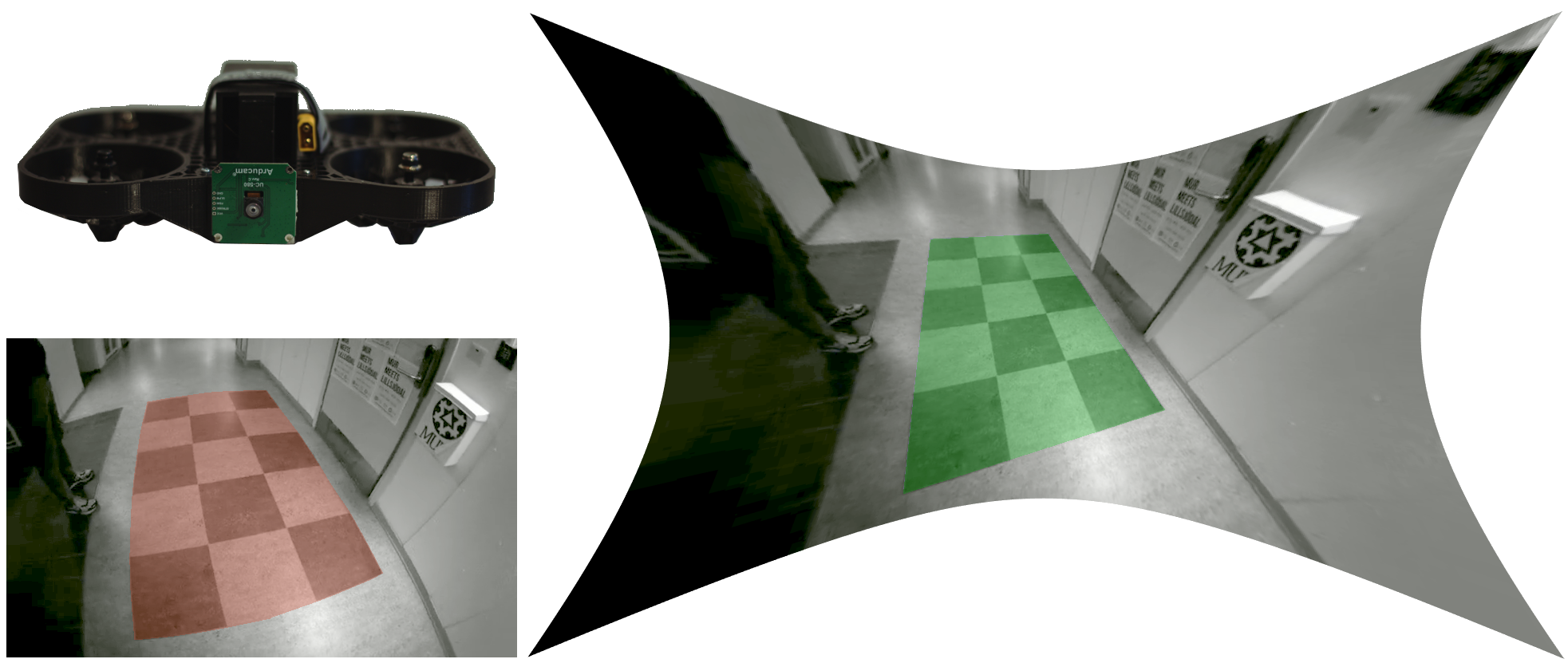}
\caption{(\emph{Left}:) The mid-sized UAV used in the dataset~\cite{valtonen-ornhag-etal-icpr-2021} and one of the input images from the \emph{Indoor} sequence. (\emph{Right}:) The output of the proposed 4-point algorithm, where
the distortion parameter was estimated using histogram voting. Note that the checkerboard pattern
on the floor is a quadrilateral box in real-life; however, it is significantly distorted (red) in the input image. The rectified image, on the other hand, displays a quadrilateral box (green).
This indicates that
lines are mapped to lines and that the pinhole camera model is applicable.
This is strong evidence that the distortion profile
has been accurately estimated, and that the one-parameter division model~\eqref{eq:divisionmodel}
is sufficient for the optics used on the UAV.
}
\hackspace{-1mm}
\label{fig:rectified_large}
\end{figure*}

In addition, we have recorded
a new dataset with a different UAV. The scenes are typically demanding,
\eg{}~an indoor sequence containing forward motion in a corridor, which is known to be hard.
We show example images from the scenes
in~\cref{fig:thumbnails}.
This dataset uses a lightweight (27 g, $92\times 92\times 29$ mm) and low-cost nano quadcopter
available under the name~\emph{Crazyflie~2.0}, captured in $640\times 480$ resolution with an RGB camera (RunCam Nano).
The ground truth was obtained using a complete SLAM system, where the reprojection error and
IMU error were jointly minimized over both camera parameters and scene points, to create a globally
consistent solution in metric scale.
In~\cref{fig:front} we show an image of the~\emph{Crazyflie~2.0}, as well
as the output from the proposed 4-point algorithm.

The main reason to use different UAVs with different components is to show the versatility
and robustness of the proposed solvers, as different setups might perform differently depending on
which IMU filtering technique or pre-integration method is used.
Furthermore, there are cases when the distortion profile of the camera optics may not be accurately
approximated by the one-parameter division model.

In the experiments we use a simple pre-integration technique~\cite{pre_integration} to obtain the estimated relative rotation.
As discussed
in~\cref{sec:whyignore}, the direction of the gravity may drift, hence the
4-point methods by Ding~\etal{}~\cite{ding-etal-cvpr-2020} will not perform optimally, as one would have to select a
filtering technique that preserves the gravity direction. To remedy this situation, we use
the ground truth poses to align the direction for this method every frame, and
apply the estimated relative orientation. Note that this gives the exact same relative error per
input and frame as for the other solvers, but guarantees that the input to the competing 4-point solver is aligned with the gravity
direction. Although these circumstances are not plausible in real-life using pre-integration,
this assures that the
result is not skewed by the choice of IMU filtering technique. The reported statistics
for the 4-point method~\cite{ding-etal-cvpr-2020}, however, are optimistic, as
as the first input orientation is perfectly aligned in the experiments. In real-life situations
they might perform worse.

Another important aspect to note is that we use consecutive frames, as we want to minimize potential
drift. This means that the translation between frames might not be very long---typically not more
than 30 cm, causing the baseline to be short. In~\cite{ding-etal-cvpr-2020} the authors only
used frames $\{\{1,\,11\},\{2,\,12\},\ldots\}$ to avoid this situation; however, we want to utilize this data,
as it is important for real-life applications concerning UAV positioning, \eg{}~moving obstacle
avoidance.

We use the following error metrics to measure the quality of the pose reconstruction
\begin{align}
    e_{\mat{R}} &= \arccos\!\left(\fr{\tr(\mat{R}_{\mathrm{GT}} \mat{R}_{\mathrm{est}}^\T) - 1}{2}\right), \\
    e_{\mat{t}} &= \arccos\!\left(\fr{\mat{t}_{\mathrm{GT}}^\T\mat{t}_{\mathrm{est}} }{\norm{\mat{t}_{\mathrm{GT}}} \norm{\mat{t}_{\mathrm{est}}}}\right), \\
    e_{f} &= \fr{|f_{\mathrm{GT}} - f_{\mathrm{est}}|}{f_{\mathrm{GT}}},
\end{align}
which have been used in a number of
works~\cite{saurer-etal-2017,ding-etal-2019-iccv,ding-etal-cvpr-2020,valtonen-ornhag-etal-icpr-2021,valtonen-ornhag-etal-wacv-2021}. The distortion profile is not as simple to measure, and we will
discuss this in~\cref{sec:real-data-raw-input}.

\begin{table*}[t]
\centering
\caption{Pose estimation error for the two datasets with rectified input images to all but the proposed 4-point method. Note that the method by~\cite{ding-etal-cvpr-2020} is given the first ground truth rotation matrix as input to ensure the assumed alignment with the $y$-axis. Therefore,
it is likely to perform worse in real-life.}

\setlength\tabcolsep{0.15cm}
{\footnotesize
\begin{tabular}{ll | rrrr | rrrr | rrrr | rrrr }
&  & \multicolumn{4}{c|}{Basement} & \multicolumn{4}{c|}{Carpet} & \multicolumn{4}{c|}{Indoor} & \multicolumn{4}{c}{Outdoor} \\
&  & Our (3 pt) & Ding & Kukelova & Our (4 pt) & Our (3 pt) & Ding & Kukelova & Our (4 pt) & Our (3 pt) & Ding & Kukelova & Our (4 pt) & Our (3 pt) & Ding & Kukelova & Our (4 pt) \\
 \toprule
                                       & Mean   & 0.435 & \textbf{0.277} & 1.336 & 0.435 & 0.628 & \textbf{0.455} & 5.315 & 0.628 & 0.607 & \textbf{0.329} & 5.582 & 0.607 & 0.423 & \textbf{0.325} & 4.391 & 0.423  \\
\multirow{-2}{*}{Rot. Error (deg.)}    & Median & 0.378 & \textbf{0.233} & 0.765 & 0.378 & 0.566 & \textbf{0.345} & 2.197 & 0.566 & 0.496 & \textbf{0.308} & 2.169 & 0.496 & 0.345 & \textbf{0.284} & 1.090 & 0.345  \\
                                       & Mean   & 5.236 & \textbf{4.525} & 11.649 & 6.127 & 3.296 & \textbf{3.129} & 11.450 & 4.871 & 4.889 & \textbf{4.179} & 20.648 & 6.093 & \textbf{5.345} & 7.138 & 25.872 & 6.446  \\
\multirow{-2}{*}{Trans. Error (deg.)}  & Median & 3.755 & \textbf{2.741} & 7.457 & 4.458 & 2.243 & \textbf{1.732} & 8.414 & 2.929 & 2.689 & \textbf{2.286} & 14.811 & 2.722 & \textbf{3.576} & 4.002 & 17.090 & 4.499  \\
                                       & Mean   & 24.804 & \textbf{20.977} & 72.814 & 27.315 & 6.921 & \textbf{6.462} & 63.394 & 46.546 & \textbf{7.595} & 9.011 & 69.189 & 16.190 & \textbf{12.914} & 27.351 & 88.031 & 39.547  \\
\multirow{-2}{*}{Focal. Error (perc.)} & Median & 6.336 & \textbf{4.887} & 18.681 & 12.293 & 3.597 & \textbf{2.526} & 23.606 & 7.761 & 4.582 & \textbf{3.322} & 39.160 & 7.546 & \textbf{7.223} & 8.923 & 39.310 & 9.584  \\
\midrule
&  & \multicolumn{4}{c|}{Bicycle lane} & \multicolumn{4}{c|}{Building} & \multicolumn{4}{c|}{Corridor} & \multicolumn{4}{c}{Office} \\
&  & Our (3 pt) & Ding & Kukelova & Our (4 pt) & Our (3 pt) & Ding & Kukelova & Our (4 pt) & Our (3 pt) & Ding & Kukelova & Our (4 pt) & Our (3 pt) & Ding & Kukelova & Our (4 pt) \\
 \toprule
                                       & Mean   & \textbf{0.247} & 1.872 & 3.888 & \textbf{0.247} & \textbf{0.223} & 0.861 & 2.016 & \textbf{0.223} & \textbf{0.181} & 0.696 & 1.644 & \textbf{0.181} & \textbf{0.201} & 0.871 & 2.935 & \textbf{0.201}  \\
\multirow{-2}{*}{Rot. Error (deg.)}    & Median & \textbf{0.197} & 0.684 & 1.658 & \textbf{0.197} & \textbf{0.197} & 0.413 & 0.902 & \textbf{0.197} & \textbf{0.143} & 0.312 & 0.866 & \textbf{0.143} & \textbf{0.167} & 0.576 & 1.398 & \textbf{0.167}  \\
                                       & Mean   & \textbf{13.863} & 23.916 & 27.650 & 19.128 & \textbf{9.534} & 15.077 & 24.944 & 15.181 & \textbf{5.403} & 10.116 & 17.330 & 7.638 & \textbf{6.250} & 10.091 & 20.422 & 15.616  \\
\multirow{-2}{*}{Trans. Error (deg.)}  & Median & \textbf{8.159} & 13.020 & 21.000 & 10.932 & \textbf{7.020} & 9.296 & 16.750 & 9.526 & \textbf{3.493} & 5.499 & 11.027 & 4.000 & \textbf{3.535} & 6.465 & 13.490 & 8.326  \\
                                       & Mean   & 32.138 & 324.201 & 54.166 & \textbf{29.940} & \textbf{14.759} & 27.875 & 54.659 & 68.120 & 34.890 & 43.671 & 64.270 & \textbf{28.590} & \textbf{14.140} & 19.288 & 43.134 & 58.680  \\
\multirow{-2}{*}{Focal. Error (perc.)} & Median & \textbf{6.104} & 10.567 & 22.768 & 10.679 & \textbf{7.979} & 10.456 & 21.067 & 13.702 & \textbf{8.382} & 10.960 & 27.977 & 11.068 & \textbf{5.935} & 6.024 & 15.369 & 17.952  \\

\end{tabular}
}

\hackspace{-3mm} 
\label{tab:real}
\end{table*}
\begin{table*}[t]
\centering
\caption{Pose estimation error with unrectified input images to all methods. The proposed 4-point method is the only method
capable of correcting for radial distortion artifacts.}

\setlength\tabcolsep{0.15cm}
{\footnotesize
\begin{tabular}{ll | rrrr | rrrr | rrrr | rrrr }
&  & \multicolumn{4}{c|}{Basement} & \multicolumn{4}{c|}{Carpet} & \multicolumn{4}{c|}{Indoor} & \multicolumn{4}{c}{Outdoor} \\
&  & Our (3 pt) & Ding & Kukelova & Our (4 pt) & Our (3 pt) & Ding & Kukelova & Our (4 pt) & Our (3 pt) & Ding & Kukelova & Our (4 pt) & Our (3 pt) & Ding & Kukelova & Our (4 pt) \\
\toprule
                                       & Mean   & \textbf{0.435} & 1.117 & 3.750 & \textbf{0.435} & \textbf{0.628} & 0.989 & 5.445 & \textbf{0.628} & \textbf{0.607} & 0.798 & 6.555 & \textbf{0.607} & \textbf{0.423} & 1.042 & 3.546 & \textbf{0.423}  \\
\multirow{-2}{*}{Rot. Error (deg.)}    & Median & \textbf{0.378} & 0.669 & 2.782 & \textbf{0.378} & \textbf{0.566} & 0.729 & 2.828 & \textbf{0.566} & \textbf{0.496} & 0.542 & 5.428 & \textbf{0.496} & \textbf{0.345} & 0.635 & 2.507 & \textbf{0.345}  \\
                                       & Mean   & 25.336 & 29.364 & 20.061 & \textbf{6.127} & 9.248 & 9.437 & 17.567 & \textbf{4.871} & 18.714 & 20.331 & 24.143 & \textbf{6.093} & 26.498 & 28.659 & 28.739 & \textbf{6.446}  \\
\multirow{-2}{*}{Trans. Error (deg.)}  & Median & 16.546 & 22.420 & 14.148 & \textbf{4.458} & 5.443 & 5.049 & 13.328 & \textbf{2.929} & 9.104 & 11.134 & 18.962 & \textbf{2.722} & 14.735 & 21.101 & 22.681 & \textbf{4.499}  \\
                                       & Mean   & 2670.475 & 735.586 & 120.351 & \textbf{27.315} & 65.415 & 67.152 & 101.243 & \textbf{46.546} & 409.987 & 231.913 & 70.609 & \textbf{16.190} & 2659.021 & 718.710 & 123.999 & \textbf{39.547}  \\
\multirow{-2}{*}{Focal. Error (perc.)} & Median & 48.684 & 94.448 & 43.793 & \textbf{12.293} & 62.921 & 56.633 & 50.641 & \textbf{7.761} & 37.991 & 62.136 & 43.290 & \textbf{7.546} & 60.240 & 129.404 & 48.381 & \textbf{9.584}  \\
\midrule
&  & \multicolumn{4}{c|}{Bicycle lane} & \multicolumn{4}{c|}{Building} & \multicolumn{4}{c|}{Corridor} & \multicolumn{4}{c}{Office} \\
&  & Our (3 pt) & Ding & Kukelova & Our (4 pt) & Our (3 pt) & Ding & Kukelova & Our (4 pt) & Our (3 pt) & Ding & Kukelova & Our (4 pt) & Our (3 pt) & Ding & Kukelova & Our (4 pt) \\
 \toprule
                                       & Mean   & \textbf{0.247} & 4.175 & 6.883 & \textbf{0.247} & \textbf{0.223} & 4.263 & 3.852 & \textbf{0.223} & \textbf{0.181} & 1.674 & 3.221 & \textbf{0.181} & \textbf{0.201} & 2.377 & 6.975 & \textbf{0.201}  \\
\multirow{-2}{*}{Rot. Error (deg.)}    & Median & \textbf{0.197} & 2.205 & 4.280 & \textbf{0.197} & \textbf{0.197} & 2.137 & 2.738 & \textbf{0.197} & \textbf{0.143} & 0.795 & 1.809 & \textbf{0.143} & \textbf{0.167} & 1.065 & 6.048 & \textbf{0.167}  \\
                                       & Mean   & 45.992 & 42.220 & 38.641 & \textbf{19.128} & 37.619 & 38.630 & 36.473 & \textbf{15.181} & 17.766 & 23.799 & 24.081 & \textbf{7.638} & 32.616 & 33.743 & 27.831 & \textbf{15.616}  \\
\multirow{-2}{*}{Trans. Error (deg.)}  & Median & 47.896 & 37.698 & 32.783 & \textbf{10.932} & 28.936 & 38.025 & 30.757 & \textbf{9.526} & 10.920 & 15.564 & 17.215 & \textbf{4.000} & 22.109 & 23.946 & 23.512 & \textbf{8.326}  \\
                                       & Mean   & 1810.425 & 620.434 & 117.463 & \textbf{29.940} & 7189.648 & 455.162 & 80.915 & \textbf{68.120} & 480.445 & 185.039 & 130.248 & \textbf{28.590} & 4293.272 & 1071.105 & 99.617 & \textbf{58.680}  \\
\multirow{-2}{*}{Focal. Error (perc.)} & Median & 250.336 & 204.291 & 44.099 & \textbf{10.679} & 44.812 & 94.533 & 43.344 & \textbf{13.702} & 39.880 & 58.036 & 51.505 & \textbf{11.068} & 333.428 & 366.350 & 54.794 & \textbf{17.952}  \\
\end{tabular}
}

\hackspace{-2mm} 
\label{tab:real_rad}
\end{table*}

\subsection{Pose Estimation with Rectified Images}\label{sec:rectimages}
In this section, we use the rectified images for the methods that do not compensate for
radial distortion (the proposed 3-point solver, the 4-point solver~\cite{ding-etal-cvpr-2020}
and the 6-point solver~\cite{kukelova-etal-2017-cvpr}), while the
proposed 4-point solver is given the distorted images as input.
We do not use the~7-point solver~\cite{jiang-etal-accv-2014} in this comparison, since it is not
a feasible competitor in real situations due to its computational complexity.
Each method is given 1,000 RANSAC iterations with the same reprojection threshold, and no extra
local optimization techniques are applied.
The results are shown in~\cref{tab:real}.

From the result, we note that there is a slight advantage in terms of accuracy
in favor of the method by Ding~\etal{}~\cite{ding-etal-cvpr-2020} compared to the proposed 3-point solver
for the dataset from~\cite{valtonen-ornhag-etal-icpr-2021} generated with the mid-sized UAV;
however, the roles are reversed when looking at the new dataset with the \emph{Crazyflie~2.0} UAV.
Note that the rotation error for the proposed 3-point (and 4-point, since they are identical)
are generally larger for the mid-sized UAV (in the range 0.4--0.6 degrees) whereas the error input to the
\emph{Crazyflie~2.0} is slightly smaller (roughly 0.2 degrees on average). This could be explained by the internal calibration
between the IMU and the camera being more accurate on the \emph{Crazyflie~2.0}.
Furthermore, we would like to emphasize that the difference between the proposed 3-point and competing 4-point method~\cite{ding-etal-cvpr-2020} is
not very large even for the mid-sized UAV and that in a real scenario one might want to choose the proposed method, as it is $800\times$ faster.
Another interesting aspect is that the visual-only method by~\cite{kukelova-etal-2017-cvpr} is performing significantly worse than the competing methods,
which was also noted in~\cite{ding-etal-cvpr-2020}.

\subsection{Pose Estimation with Raw Images}\label{sec:real-data-raw-input}
We now turn our attention to using distorted input images. This scenario is interesting for UAV operators who wish to change optics out in the field
without intermediate calibration procedures. The same input sequences as in~\cref{sec:rectimages} are used; however, they are not rectified prior to
estimating the image point correspondences. The results are shown in~\cref{tab:real_rad}. Unsurprisingly, our 4-point method outperforms the other methods
that cannot correct for distortion artifacts. What is perhaps more interesting is that the performance, in general, is better than the visual-only 6-point method~\cite{kukelova-etal-2017-cvpr} on rectified data. These observations suggest
that the radial distortion auto-calibration approach is practically feasible using the proposed solver.

As optics, in general, are not perfectly approximated by the one-parameter division model, it is non-trivial to express the performance of the radial distortion correction.
Instead, we
rely on an ocular inspection of the estimated radial distortion parameter for two sequences. In~\cref{fig:front} and~\cref{fig:rectified_large} we show
the rectifications, using the estimated radial distortion parameter obtained from histogram voting of the respective sequence. In the latter case, we get a clear visual confirmation of the successful estimation of the
radial distortion parameter, in the form of a quadrilateral checkerboard pattern visible on the floor.

\section{Conclusions}
In this paper, we have investigated an assumption of ignoring the IMU drift for short time intervals.
We showed that modern pre-integration methods perform well and that the relative pose problem can
be solved accurately and satisfactorily using this assumption. What is most important, is that the
resulting equations are significantly easier to solve, opening up the possibility to tackle
problems that were previously considered extremely hard and not suitable for real-time applications.
We proposed the first-ever minimal solver for simultaneously estimating the focal length,
distortion profile, and motion parameters while incorporating the IMU data. Furthermore, we showed
a speed-up of~$800\times$ compared to the current state-of-the-art for the partially calibrated
case with unknown and equal focal length, with little to no loss in accuracy. The methods have been
thoroughly tested on different UAVs with different components, in several challenging indoor
and outdoor environments, demonstrating excellent performance.

\clearpage

{\small
    \bibliographystyle{ieee_fullname}
    \bibliography{uav_essential}
}

\end{document}